\documentclass[11pt]{article}

\usepackage[final]{acl} 

\usepackage{times}
\usepackage{latexsym}
\usepackage[T1]{fontenc}
\usepackage[utf8]{inputenc}
\usepackage{microtype}
\usepackage{inconsolata}

\usepackage{graphicx} 
\usepackage{booktabs} 
\usepackage{multirow} 
\usepackage{amsmath}  
\usepackage{amssymb} 
\usepackage{tabularx} 
\usepackage{enumitem} 
\usepackage{siunitx} 
\usepackage{makecell} 
\usepackage{natbib}
\usepackage{float}

\newcolumntype{T}[1]{S[table-format=#1]}

\title{Automated Multiple Mini Interview (MMI) Scoring} 

\author{Ryan Huynh \\
  University of Surrey\\
  \texttt{rh00902@surrey.ac.uk} \\\And
  Frank Guerin \\
  University of Surrey \\
  \texttt{f.guerin@surrey.ac.uk} \\\And
  Alison Callwood \\
  University of Surrey \\
  \texttt{a.callwood@surrey.ac.uk} \\}

\begin{document}
\maketitle

\begin{abstract}
Assessing soft skills such as empathy, ethical judgment, and communication is essential in competitive selection processes, yet human scoring is often inconsistent and biased. While Large Language Models (LLMs) have improved Automated Essay Scoring (AES), we show that state-of-the-art rationale-based fine-tuning methods struggle with the abstract, context-dependent nature of Multiple Mini-Interviews (MMIs), missing the implicit signals embedded in candidate narratives. We introduce a multi-agent prompting framework that breaks down the evaluation process into transcript refinement and criterion-specific scoring. Using 3-shot in-context learning with a large instruct-tuned model, our approach outperforms specialised fine-tuned baselines (Avg QWK 0.62 vs 0.32) and achieves reliability comparable to human experts. We further demonstrate the generalisability of our framework on the ASAP benchmark, where it rivals domain-specific state-of-the-art models without additional training. These findings suggest that for complex, subjective reasoning tasks, structured prompt engineering may offer a scalable alternative to data-intensive fine-tuning, altering how LLMs can be applied to automated assessment.
\end{abstract}


\section{Introduction}
\label{sec:intro}
Assessing critical soft skills is a fundamental challenge in competitive selection processes. Skills such as empathy, communication, and ethical judgment are often stronger predictors of professional success than technical knowledge alone, particularly in fields like healthcare \citep{eva2004admissions}. The Multiple Mini-Interview (MMI) was designed specifically to measure these abilities and mitigate the biases of traditional interviews. However, a significant challenge persists: inconsistent scoring by human interviewers, stemming from fatigue and subjective judgment, can undermine the fairness and reliability the MMI seeks to achieve \citep{humphrey2008multiple, roberts2025professional}. Automating this scoring process therefore presents a critical opportunity to enhance consistency and ensure equitable assessment.

Recent advancements in Large Language Models (LLMs) present a promising opportunity, but automating MMI scoring is a far more complex challenge than tasks like Automated Essay Scoring (AES) \cite{mansour2024can}. The primary challenge is not the assessment of explicit text, but the identification of abstract, context-dependent soft skills, which are embedded indirectly within a candidate's narrative. A single phrase or action can signify empathy or not based entirely on context. This is the core difficulty, which we showed in our initial analysis (Section \ref{sec:methodology:initial}), as simpler, embedding-based models fail to distinguish between semantically similar but contextually opposite actions (e.g., walking towards a grieving friend versus away from them).

Given the success of AI in AES, a logical first step is to adapt state-of-the-art (SOTA) frameworks from that domain. For instance, the rationale-based fine-tuning approach of \citet{chu2025rationale} has shown SOTA results in AES on the ASAP dataset, exceeding human-to-human correlations. However, we found this method does not transfer effectively to our MMI dataset. One major cause lies in the differences between the scoring rubrics. The ASAP dataset benefits from detailed, prompt-specific rubrics for every trait. In contrast, the MMI rubric describes broad, abstract interpersonal qualities, such as empathy, that often measure differently depending on context across scenarios. This means there are multiple valid interpretations of what makes up a “strong” response. This failure of existing SOTA methods on our task highlights a clear research gap and necessitates an approach tailored to the interpretive nature of MMI scoring.

We develop and validate a methodology for the automated scoring of soft skills. We demonstrate that a simple multi-agent prompting framework, which breaks down the evaluation into a series of focused, single-criterion scoring tasks, yields the most accurate and reliable results (Section \ref{sec:methodology:proposed}). This framework, which combines a preprocessing agent with 3-shot in-context learning, achieves high agreement with human experts when implemented with the SOTA model (Section \ref{sec:results:sota}).

We further demonstrate the generalisability of our framework on the ASAP dataset. Whilst we show the SOTA AES method \cite{chu2025rationale} fails on our complex MMI dataset, our MMI-derived framework is comparable on the essay-type questions of the ASAP benchmark dataset (Section \ref{sec:results:asap}). When applied to ASAP, our method performs on par with human-to-human reliability, demonstrating the potential as a transferable methodology not just specific for MMI but, also for multi-trait automated assessment.

The main contributions of this paper are as follows:
\begin{itemize}[noitemsep,topsep=0pt] 
    
    \item We demonstrate that a simple multi-agent prompting framework, which breaks down the assessment into criterion-specific 3-shot scoring outperforms SOTA fine-tuning approaches on soft-skill assessments in Multiple Mini-Interviews.
    \item We illustrate the generalisability of our framework by applying it to the widely-used ASAP dataset, where it achieves performance competitive with specialised, SOTA models. 
\end{itemize}

\section{Related Work}
\label{sec:related_work}



The most comparable domain to our task is Automated Essay Scoring (AES), where systems are designed to automatically score written text. Early AES systems relied on handcrafted linguistic features and traditional machine-learning models \citep{attali2006automated}. The field advanced with deep learning and transformer-based models such as BERT \citep{mayfield-black-2020-fine, wang-etal-2022-use}, while public benchmarks like the Automated Student Assessment Prize (ASAP) dataset\footnote{\url{https://www.kaggle.com/datasets/lburleigh/asap-2-0}} accelerated progress.

Recent work uses zero- and few-shot prompting with LLMs such as GPT-3.5 and Llama, achieving moderate results, though still below human-level agreement \citep{kundu2024are, tang-chen-lin-2024}. \citet{do-kim-lee-2024-autoregressive} proposed an autoregressive scoring model (ArTS) using T5 for multi-trait prediction, improving both correlation and reliability. The state-of-the-art work by \citet{chu2025rationale} introduced the Rationale-based Multiple Trait Scoring (RMTS) framework, where a larger LLM generates trait-specific rationales used by a smaller fine-tuned model. Other studies integrate linguistic features into LLM-based AES \citep{hou-ciuba-li-2025-improving} or develop zero-shot systems that specialise to traits without supervision \citep{lee-etal-2024-unleashing}. 

While fine-tuning approaches tend to achieve higher accuracy, they require significant training sets (ASAP has 15k), which might be difficult to obtain when, e.g.,  new interview questions are designed for each year. 
An alternative is to instead rely on few-shot prompting, with a small number of scored examples in the prompt \citep{yoshida2024impact}.
This draws on the results of  earlier works on in-context learning, exploring how sensitive it is to exemplar choice and different selection strategies  \citep{zhang2022active, liu2022makes}. Our study directly investigates exemplar selection and prompt configuration for a subjective scoring task to identify the most effective, low-cost setup.

The scoring rubric is a key component of any automated assessment task. 
Recent work in LLM-based scoring has shown that performance is linked to the quality and detail of the  rubric 
\citep{yavuz2025utilizing, hashemi-etal-2024-llm, pathak-etal-2025-rubric}.  
Standard AES tasks, such as those using the ASAP dataset, benefit from detailed, concrete, question-specific rubrics that define traits \citep{chu2025rationale}. 
This contrasts with our MMI scoring, where we only have a  single, broad rubric that must be applied across many different scenarios. 



\section{The MMI Scoring Task and Dataset}
\label{sec:task}

The task is defined as follows: given a transcribed MMI response,  predict a score on a 7-point Likert scale for each of the nine distinct criteria.
To measure the agreement between the model's predictions and the human expert scores, we use two standard metrics for inter-rater reliability. The primary metric is Quadratic Weighted Kappa (QWK), which measures agreement on an ordinal scale while penalising larger differences between scores more heavily. It is the standard metric in automated assessment literature. We also report Mean Squared Error (MSE) to provide a direct measure of the average magnitude of the scoring error.

The data for this study was collected from a virtual MMI (VMMI) system used for university admissions in healthcare programs. The dataset consists of 1001 candidate responses to four distinct scenario-based questions. To mitigate potential biases from appearance, voice, or accent, our analysis uses only automatically transcribed text from AWS transcription services (2022-2023).

Each response was evaluated by human expert assessors on a 7-point Likert scale (where 1 is "Unsatisfactory" and 7 is "Excellent") across a set of soft-skill criteria. Since our analysis is text-based, we excluded criterion c1 which focused on non-verbal communications. The nine textual criteria we evaluate are named as c2 to c10 (“c” represents criteria)\footnote{Due to ethical considerations regarding confidentiality and commercial conflicts. This dataset and criteria descriptions cannot be shared publicly but are available on reasonable request.}.

As the exact scenario-based questions cannot be shared, we provide a brief summary. The four scenarios involved candidates describing their response to: consoling a grieving friend (Q3), facing professional inadequacy (Q4), navigating a public ethical dilemma (Q5), and resolving team conflict (Q6). A more detailed description of each scenario is available in Appendix \ref{sec:appendix:scenarios}.

\begin{figure}[t!]
\centering
\includegraphics[width=\columnwidth]{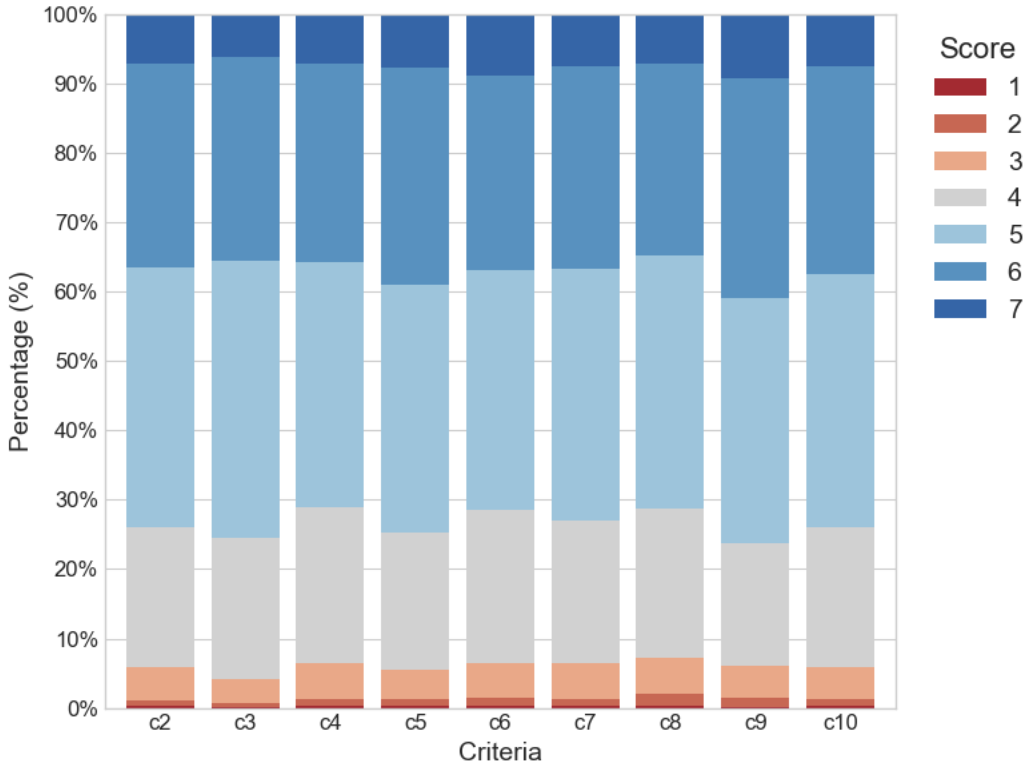}
\caption{Distribution of human-assigned scores across all criteria, showing a clear skew towards the upper end.}
\label{fig:score_dist}
\end{figure}

An initial analysis of the dataset reveals a distribution leaning towards higher scores, with 85\% of human-assigned ratings being either 4, 5 or 6 (see Fig.~\ref{fig:score_dist}). The lack of low scores was predictable given that candidates had already passed an initial screening. This concentration of competent performances makes the task of differentiating between fine-grained levels of performance particularly challenging. Further descriptive statistics 
are available in Appendix \ref{sec:appendix:scenarios} (Table \ref{tab:descriptive_stats}).

\section{Methodology}
\label{sec:methodology}


\subsection{Initial Study: Sentence-Level Analysis}
\label{sec:methodology:initial}

Building on established AES methodologies that utilise BERT-based representations to capture semantic features \citep{mayfield-black-2020-fine, wang-etal-2022-use}, our initial approach investigated whether sentence-level representations could differentiate between responses of varying quality. We used Sentence-BERT to generate embeddings for each sentence in the MMI transcripts and applied K-means clustering.

An analysis of the clusters highlighted meaningful insights into how this method captured thematic groupings. For example, the smallest cluster (Cluster 45) primarily contained filler words like "mmm" or "thank you," which lack evaluative content. In contrast, the largest clusters revealed recurring themes. For Question 3 (consoling a friend), Cluster 20 predominantly consisted of sentences expressing empathy or understanding. For Question 5 (public dilemma), Cluster 31 grouped sentences related to behavioural actions, such as "talk softly" or "ignore the rude people."

However, this thematic grouping proved superficial for scoring. We found that sentences with largely different scoring implications were often grouped together. For instance, one cluster contained sentences thematically related to "distance" (e.g., "moving toward the friend" vs. "walking away from the friend"). While semantically similar, these actions have opposite values in an assessment of empathy. This key failure highlighted the limitations of this approach: while the embeddings successfully captured semantic concepts, they prioritised thematic similarity over evaluative alignment with the rubric. As a result, the clusters grouped sentences based on the nature of the action described rather than its appropriateness or effectiveness in the given scenario. For instance, distinct actions were often clustered together simply because they shared a semantic theme, regardless of their differing outcomes. This illustrates a broader limitation of embedding-based AES methods, where semantic similarity fails to distinguish between positive and negative behaviours when viewed through the lens of the specific soft-skill criteria. This confirmed that a more advanced model, capable of interpreting the implications of an action relative to the scoring standards, was necessary.




\subsection{Iterative Development of Prompting Strategies}
\label{sec:methodology:prompting}

For our primary development, we used instruct-tuned versions of the Llama model family (8B, 70B, and 405B parameters). A standardised prompt template was created to serve as a consistent base for all experiments. The text for this base template, along with the final, optimal templates, is provided in Appendix \ref{sec:appendix:prompts}.

We conducted a series of experiments to identify the most effective prompting technique (full results for all configurations are detailed in Appendix \ref{sec:appendix:prompt_rag_results}).

\subsubsection{In-Context Learning}
\label{sec:methodology:context_learning}
We began by comparing zero-shot and few-shot prompting across all three Llama model sizes. While performance generally scaled with model size, the 405B model consistently performed best.

An important observation across all models was a tendency to overestimate scores. This inherent bias meant that the choice of in-context examples was a critical factor in calibration. Examples were carefully selected from the 5th (low), 50th (medium), and 95th (high) percentiles of the training data to ensure they were representative. We found that providing examples from only one extreme (whether only high or only low) resulted in poorer performance. While high-scoring examples reinforced the overestimation bias, low-scoring examples failed to define the upper bounds of the scoring scale, leaving the model unable to distinguish higher-quality responses. A balanced set was essential, as having low, medium and high examples provided a necessary reference point.

Performance improved with the number of examples up to a clear threshold. For Question 3, the average QWK improved from 0.215 (zero-shot) to 0.363 with a 3-shot (low/medium/high) configuration. We tested larger contexts (4-shot and 5-shot), but performance deteriorated. We observed that with four or more examples, the model's prediction became heavily biased towards the score of the final example in the prompt. This suggested that three examples represented an effective limit for the model to generalise from. Thus, the 3-shot (L/M/H) strategy was identified as our optimal single-prompt configuration.

\subsubsection{Retrieval-Augmented Generation (RAG)}
\label{sec:methodology:RAG}
Next, we hypothesised retrieving the most semantically similar responses as in-context examples could provide more tailored context. We implemented an embedding-based retrieval system (implementation details provided in Appendix \ref{sec:appendix:rag_details}) and tested several RAG strategies. First, retrieving the three most similar responses performed poorly, as the dataset's imbalance (majority of high scores) resulted in a retrieved set that lacked the necessary score diversity for calibration. Second, we combined our optimal 3-shot (L/M/H) base examples with one additional retrieved example. This approach also performed poorly, as it replicated the issues from our 4-shot prompting tests, where the addition of a fourth example introduced noise rather than improving generalisation. Finally, we tested an adaptive approach where we retrieved a single most similar response and used it to replace only the specific base example (Low, Medium, or High) that corresponded to its score tier, while keeping the other two static examples fixed. This final strategy was the best-performing RAG method, achieving a QWK of 0.329 on Question 3.

However, as none of these RAG strategies surpassed the performance of our fixed 3-shot strategy (QWK 0.3630), we concluded that for this specific task, where calibration across a skewed score distribution is critical, a carefully balanced, static set of examples proved more effective than dynamic retrieval.

\subsection{Proposed Method: A Multi-Agent Framework}
\label{sec:methodology:proposed}

Our final and best-performing method was developed by building upon our findings. We hypothesised that tasking a single prompt with evaluating nine distinct soft skills simultaneously induced cross-criterion interference, a limitation observed in recent multi-trait scoring research \citep{lee-etal-2024-unleashing}. To address this, we designed a two-stage, multi-agent framework that enforces independent evaluation through trait-specific reasoning. This approach yielded a QWK of 0.533 on Question 3, a substantial improvement over the 0.363 achieved by the best single-prompt configuration, establishing it as our optimal methodology. The framework consists of two sequential stages:
\begin{enumerate}[noitemsep,topsep=0pt]
    \item \textbf{The Preprocessing Agent:} The raw VMMI transcripts contained conversational filler and other noise. This agent's sole task was to refine the raw transcript, producing a concise version of the candidate's core response.
    \item \textbf{The Scoring Agents:} Following preprocessing, the task was divided among nine separate agents, one for each criterion. Each agent was given a tailored prompt that included a base 7-point scoring rubric and utilised our optimal 3-shot (L/M/H) strategy. Crucially, the examples provided to each agent were selected based on their percentile ranking for that specific criterion, ensuring maximum relevance to the skill being evaluated.
\end{enumerate}

\subsection{Fine-Tuning vs. Prompting}
\label{sec:results:finetuning}

We conducted a series of fine-tuning experiments to determine if a smaller, specialised model could match or exceed the performance of our prompt-engineered framework.

\subsubsection{Baseline Fine-Tuning}
\label{sec:results:finetuning_baseline}
We selected two distinct model architectures for this task. First, we selected Llama 3.1 8B, an open-source decoder-based model. For this model, we employed a supervised generative fine-tuning approach, where the model was trained to output the numerical score directly as text tokens. Second, we chose modernBERT, a state-of-the-art BERT-based model \citep{warner2024modernbert}. This encoder-only model served as a logical successor to our initial SentenceBERT analysis (Section \ref{sec:methodology:initial}), representing a more traditional fine-tuning approach for classification and regression tasks.

Full implementation details of the fine-tuning setup, including data splits, optimisation parameters, quantisation, and LoRA configuration, are provided in Appendix \ref{sec:appendix:finetune_setup}.

Using this setup, we tested several training strategies inspired by our prompting experiments: a Grouped approach, where a single model was trained to score all criteria from one prompt; an Individual approach, where separate models were trained for each specific criterion; and a Preprocessing strategy, which used the cleaned transcripts from our preprocessing agent (Section \ref{sec:methodology:proposed}) as input. The Llama 3.1 8B model was trained under all four conditions (Grouped, Individual, with and without preprocessing). As modernBERT is more suited to single-task predictions, it was trained only using the "Individual" approach, with and without preprocessing. The results are shown in Table \ref{tab:finetune_baseline}.

\begin{table*}[ht!]
\centering
\small
\setlength{\tabcolsep}{2.5pt} 
\begin{tabular}{l T{1.4}T{1.4} T{1.4}T{1.4} T{1.4}T{1.4} T{1.4}T{1.4} T{1.4}T{1.4} T{1.4}T{1.4} }
\toprule
 & \multicolumn{2}{c}{Llama-G} & \multicolumn{2}{c}{Llama-I} & \multicolumn{2}{c}{mBERT} & \multicolumn{2}{c}{Llama-G-P} & \multicolumn{2}{c}{Llama-I-P} & \multicolumn{2}{c}{mBERT-P} \\
\cmidrule(lr){2-3} \cmidrule(lr){4-5} \cmidrule(lr){6-7} \cmidrule(lr){8-9} \cmidrule(lr){10-11} \cmidrule(lr){12-13}
Crit. & {QWK$^\uparrow$} & {MSE$^\downarrow$} & {QWK$^\uparrow$} & {MSE$^\downarrow$} & {QWK$^\uparrow$} & {MSE$^\downarrow$} & {QWK$^\uparrow$} & {MSE$^\downarrow$} & {QWK$^\uparrow$} & {MSE$^\downarrow$} & {QWK$^\uparrow$} & {MSE$^\downarrow$} \\
\midrule
c2 & 0.264 & 1.27 & 0.241 & 1.23 & 0.242 & 1.10 & 0.242 & 1.24 & 0.332 & 0.951 & 0.265 & 1.04 \\
c3 & 0.200 & 1.22 & 0.106 & 1.28 & 0.233 & 0.991 & 0.209 & 1.17 & 0.186 & 1.02 & 0.237 & 1.00 \\
c4 & 0.260 & 1.28 & 0.193 & 1.32 & 0.254 & 1.06 & 0.269 & 1.26 & 0.157 & 1.10 & 0.233 & 1.08 \\
c5 & 0.246 & 1.29 & 0.154 & 1.27 & 0.219 & 1.15 & 0.273 & 1.21 & 0.266 & 0.963 & 0.251 & 1.03 \\
c6 & 0.281 & 1.32 & 0.220 & 1.23 & 0.312 & 1.01 & 0.286 & 1.27 & 0.327 & 0.988 & 0.320 & 1.19 \\
c7 & 0.258 & 1.28 & 0.161 & 1.43 & 0.217 & 1.01 & 0.254 & 1.24 & 0.352 & 0.938 & 0.234 & 1.00 \\
c8 & 0.215 & 1.41 & 0.123 & 1.61 & 0.206 & 1.14 & 0.248 & 1.32 & 0.315 & 1.06 & 0.265 & 1.20 \\
c9 & 0.237 & 1.35 & 0.140 & 1.59 & 0.240 & 1.08 & 0.248 & 1.30 & 0.398 & 0.900 & 0.255 & 1.16 \\
c10& 0.263 & 1.30 & 0.189 & 1.51 & 0.286 & 1.08 & 0.286 & 1.21 & 0.515 & 0.700 & 0.251 & 1.16 \\
\midrule
\textbf{Avg} & 0.247 & 1.30 & 0.170 & 1.39 & 0.245 & 1.09 & 0.257 & 1.25 & \textbf{0.316} & \textbf{0.958} & 0.257 & 1.10 \\
\bottomrule
\end{tabular}
\caption{Performance Comparison of Baseline Fine-Tuned Models. G=Grouped, I=Individual, P=Preprocessed, mBERT=modernBERT. Llama 3.1 8B 4-bit quantized was used. Best average results in bold.}
\label{tab:finetune_baseline}
\end{table*}

The results in Table \ref{tab:finetune_baseline} align with our findings from prompt engineering. The Llama 3.1 8B - Individual, Preprocessed model achieved the highest average QWK (0.316) and lowest MSE (0.958). This further strengthens the finding that breaking down the task by criterion ("Individual") and cleaning the data ("Preprocessed"), are beneficial in a fine-tuning context. The modernBERT models showed a similar, smaller, benefit from preprocessing (Avg QWK improving from 0.245 to 0.257), but their overall performance did not reach the level of the best fine-tuned Llama model.

\subsubsection{SOTA-Adapted Fine-Tuning (RMTS)}
We further investigated whether a more sophisticated, state-of-the-art fine-tuning method could bridge the performance gap. We adapted the Rationale-based Multiple Trait Scoring (RMTS) framework from \citet{chu2025rationale}. In this method, an LLM is first used to generate rationales (justifications) for scores, which are then used as an input feature for fine-tuning a smaller model.

We adapted this process for our MMI task with several modifications. First, we used Llama 4 Maverick to generate justifications for all criteria. We expanded the prompt to generate justifications up to 150 words (from the original 50) and mandated the inclusion of direct quotations. We also prompted for a predicted score alongside the rationale. Finally, while the original RMTS paper trained only on the justifications, we fine-tuned our Llama 3.1 8B and modernBERT models on a richer input: the preprocessed candidate response + the generated justification + the justification's score. The results are shown in Table \ref{tab:finetune_rmts}.

\begin{table}[ht!] 
\centering
\small
\setlength{\tabcolsep}{5pt} 
\begin{tabular}{l T{1.4}T{1.4}@{\hspace{1em}} T{1.4}T{1.4}} 
\toprule
\multirow{3}{*}{Criterion} & \multicolumn{2}{c}{\makecell{Llama 3.1 8B \\ (P + Just.)}} & \multicolumn{2}{c}{\makecell{modernBERT \\ (P + Just.)}} \\
\cmidrule(lr){2-3} \cmidrule(lr){4-5}
 & {QWK$^\uparrow$} & {MSE$^\downarrow$} & {QWK$^\uparrow$} & {MSE$^\downarrow$} \\
\midrule
c2 & 0.270 & 1.26 & 0.285 & 1.12 \\
c3 & 0.190 & 1.24 & 0.241 & 1.05 \\
c4 & 0.295 & 1.31 & 0.315 & 1.07 \\
c5 & 0.298 & 1.26 & 0.311 & 1.02 \\
c6 & 0.335 & 1.32 & 0.345 & 1.18 \\
c7 & 0.296 & 1.32 & 0.297 & 1.13 \\
c8 & 0.277 & 1.30 & 0.337 & 1.19 \\
c9 & 0.277 & 1.30 & 0.306 & 1.13 \\
c10& 0.362 & 1.19 & 0.357 & 0.997 \\
\midrule
\textbf{Average} & 0.288 & 1.29 & \textbf{0.310} & \textbf{1.10} \\
\bottomrule
\end{tabular}
\caption{Performance of SOTA-Adapted (RMTS) Fine-Tuned Models. P=Preprocessed, Just.=Justification. Best average results in bold.}
\label{tab:finetune_rmts}
\end{table}

The results of this SOTA-adapted method were mixed. For modernBERT, the RMTS-adapted approach (Avg QWK 0.310) showed an improvement over its baseline fine-tuning (Avg QWK 0.257). Conversely, for Llama 3.1 8B, this method (Avg QWK 0.288) actually performed worse than its best baseline (Avg QWK 0.316). This suggests that while rationale-based features can be beneficial for encoder-only models, they may introduce conflicting information or noise for decoder-based models in this setup.

\subsubsection{Analysis of Fine-Tuning Failures}
\label{sec:results:analysis_finetuning}
The underperformance of the SOTA-adapted method on our MMI dataset, despite its success on the ASAP dataset, prompted further analysis. We hypothesise this failure stems principally from the granularity of the scoring rubrics. The ASAP dataset benefits from detailed, prompt-specific rubrics that provide concrete, verifiable criteria for each trait, even for more open-ended ones like 'Ideas'. In contrast, our MMI dataset utilises a broad, abstract, and static rubric applied across all scenarios. While it is certainly possible to design detailed, question-specific rubric for soft skills, it was not part of the ground truth data available for this study. Consequently, a generated rationale for 'Empathy' is not verifiable evidence (like a rationale for 'Ideas'), but rather another high-level, subjective interpretation, which we found introduced noise and inaccuracy during fine-tuning.

To support this, we conducted two error analyses. First, we checked for overfitting to the rationale's score. We found the fine-tuned models only agreed with the rationale 37–52\% of the time, confirming they were attempting to learn from features rather than merely copying the label. Second, we analysed the prediction error distribution. Unlike the prompting models, which showed overestimation bias, the fine-tuned models exhibited balanced but widely spread errors, indicating high uncertainty. The full quantitative analysis is provided in Appendix~\ref{sec:appendix:error_dist}.

In summary, our experiments reveal that while basic optimisations like preprocessing provide marginal gains, advanced fine-tuning methods struggle to adapt to the abstract nature of MMI scoring. This highlights a critical distinction that fine-tuning thrives on the verifiable signals of detailed rubrics (as in AES) but falters on the subjective criteria of soft-skill assessment. These findings suggest that for tasks lacking detailed rubrics, the reasoning flexibility of large-scale prompt engineering may offer a more effective solution.

\section{Results and Analysis}
\label{sec:results}


\subsection{Main Results: SOTA Model Comparison}
\label{sec:results:sota}

To assess the generalisability and benchmark the performance of our optimised methodology, we conducted a final experiment using the four leading commercial models: Llama 4 Maverick, GPT-5, Gemini 2.5 Pro, and DeepSeek Reasoner. The multi-agent framework (Section \ref{sec:methodology:proposed}) was applied to the full set of four scenario-based questions (Questions 3, 4, 5, and 6).

The results, summarised in Table \ref{tab:sota_summary}, show the average performance of each model across all criteria for each question, as well as a final average representing the model's overall performance.


\begin{table}[ht!]
\centering
\small
\setlength{\tabcolsep}{4pt} 
\begin{tabular}{llrrrr}
\toprule
Metric & Q & Llama 4 & GPT 5 & Gemini 2.5 & DeepSeek \\
& & Maverick & & Pro & Reasoner \\
\midrule
\multirow{5}{*}{MSE$^\downarrow$} 
& 3 & 0.711 & 0.984 & 1.11 & 0.810 \\
& 4 & 0.781 & 1.08 & 1.34 & 0.889 \\
& 5 & 1.19 & 1.69 & 1.65 & 1.36 \\
& 6 & 0.805 & 1.25 & 0.997 & 0.986 \\
\cmidrule{2-6}
& Avg & \textbf{0.871} & 1.25 & 1.28 & 1.01 \\
\midrule
\multirow{5}{*}{QWK$^\uparrow$} 
& 3 & 0.650 & 0.448 & 0.692 & 0.473 \\
& 4 & 0.688 & 0.473 & 0.552 & 0.544 \\
& 5 & 0.434 & 0.351 & 0.473 & 0.327 \\
& 6 & 0.711 & 0.458 & 0.754 & 0.520 \\
\cmidrule{2-6}
& Avg & \textbf{0.621} & 0.432 & 0.618 & 0.466 \\
\bottomrule
\end{tabular}
\caption{Comparison of SOTA Models on the MMI Scoring Task. Q=Question, Avg=Average.}
\label{tab:sota_summary}
\end{table}

The results identify Llama 4 Maverick as the top-performing model overall, although we note the prompt was optimised for this model (see \hyperref[sec:Limitations]{Limitations}). It achieved the lowest average Mean Squared Error (MSE) of 0.871, indicating its predictions had the smallest deviation from human scores. Critically, it also recorded the highest average Quadratic Weighted Kappa (QWK) of 0.621, signifying the strongest and most consistent agreement with human expert assessors. 

Google Gemini 2.5 Pro also demonstrated highly competitive performance, achieving a close average QWK of 0.618. Notably, it achieved the highest single-question QWK (0.754) on Question 6, suggesting it may excel in specific contexts. GPT-5 (Avg QWK 0.432) and DeepSeek Reasoner (Avg QWK 0.466) delivered competent results but did not consistently reach the same level of agreement as the top two models in this task.

A consistent trend observed across all models was a dip in performance for Question 5. This scenario, which involved a complex public dilemma with multiple characters, was the most intricate of the four. This performance drop may highlight a current limitation of LLMs in evaluating responses to more complex, multi-user social situations.

Operational cost is also critical for real-world deployment. Llama 4 Maverick and DeepSeek Reasoner were the most economical, whereas Gemini 2.5 Pro and GPT-5 were substantially more expensive ($15\times$ and $6\times$ the cost of Llama 4, respectively). Considering cost-effectiveness and accuracy, Llama 4 Maverick represents the optimal choice for this MMI scoring task.

A detailed, criterion-level breakdown of all scores is available in Appendix \ref{sec:appendix:sota_details}.

\subsection{Case Study: Generalisability on the ASAP Dataset}
\label{sec:results:asap}

To validate our methodology and demonstrate its transferability beyond MMI scoring, we conducted a case study on the well-established Automated Essay Scoring (AES) task. We selected the Automated Student Assessment Prize (ASAP) dataset, focusing specifically on prompts 7 and 8. These prompts were chosen not only for their narrative structure but also for their thematic alignment with soft skill assessment. Prompt 7 explores the concept of patience, while Prompt 8 examines the significance of laughter. This focus on emotional and interpersonal themes makes them the most comparable to our MMI dataset.

We applied our optimal multi-agent framework using Llama 4 Maverick with a 3-shot (Low/Medium/High) strategy. Table \ref{tab:asap_results} presents the comparative Quadratic Weighted Kappa (QWK) results against human raters and SOTA baselines.

\begin{table*}[ht!]
\centering
\small
\setlength{\tabcolsep}{3pt} 
\begin{tabular}{l l ccc ccc c c}
\toprule
Task & Model & Ideas & Organi- & Voice & Word & SF & Conven- & Style & Overall \\
& & & sation & & Choice & & tions & & \\
\midrule
\multirow{6}{*}{7} & Human Rater 1 - Human Rater 2 & 0.695 & 0.576 & - & - & - & 0.567 & 0.544 & 0.620 \\
& GPT 3.5 (1 shot) \citep{mansour2024can} & 0.045 & 0.068 & - & - & - & 0.097 & 0.079 & 0.073 \\
& Llama 2 (1 shot) \citep{mansour2024can} & 0.091 & 0.023 & - & - & - & 0.327 & 0.154 & 0.151 \\
& \textbf{Llama 4 (Ours - Base)} & \textbf{0.693} & \textbf{0.647} & - & - & - & \textbf{0.615} & \textbf{0.597} & \textbf{0.638} \\
& \textit{Llama 4 (No Examples)} & 0.601 & 0.498 & - & - & - & 0.488 & 0.445 & 0.508 \\
& \textit{Llama 4 (Reduced Rubric)} & 0.575 & 0.541 & - & - & - & 0.639 & 0.490 & 0.561 \\
\midrule
\multirow{6}{*}{8} & Human Rater 1 - Human Rater 2 & 0.531 & 0.542 & 0.467 & 0.482 & 0.507 & 0.547 & - & 0.530 \\
& GPT 3.5 (1 shot) \citep{mansour2024can} & 0.185 & 0.256 & 0.155 & 0.150 & 0.203 & 0.311 & - & 0.210 \\
& Llama 2 (1 shot) \citep{mansour2024can} & 0.276 & 0.271 & 0.278 & 0.268 & 0.121 & 0.089 & - & 0.217 \\
& \textbf{Llama 4 (Ours - Base)} & \textbf{0.694} & \textbf{0.638} & \textbf{0.687} & \textbf{0.691} & \textbf{0.621} & \textbf{0.644} & - & \textbf{0.663} \\
& \textit{Llama 4 (No Examples)} & 0.467 & 0.537 & 0.557 & 0.548 & 0.527 & 0.383 & - & 0.503 \\
& \textit{Llama 4 (Reduced Rubric)} & 0.515 & 0.450 & 0.591 & 0.556 & 0.575 & 0.573 & - & 0.543 \\
\bottomrule
\end{tabular}
\caption{Comparison of QWK scores on the ASAP dataset (Prompts 7 \& 8). We benchmark our optimal framework against human raters and baselines, alongside ablation experiments testing the necessity of few-shot examples and detailed rubrics. SF = Sentence Fluency.}
\label{tab:asap_results}
\end{table*}

Our MMI-derived framework outperformed the baseline prompting results from \citet{mansour2024can}. It also surpassed the human-to-human inter-rater reliability reported for the dataset on both prompts. When compared to the SOTA fine-tuned RMTS model \cite{chu2025rationale}: for Question 7, our average QWK (0.638) was competitive with the SOTA result (0.749); whereas for Question 8, our average QWK (0.663) was slightly higher than the SOTA result (0.658). This performance gap between the two prompts, where Prompt 8 outperforms Prompt 7, can be attributed to various differences in the dataset, such as higher grade level (Grade 10 vs 7), longer responses (average of 650 vs 250 words), and a wider scoring scale (1-6 vs 0-3). These factors suggest that our framework benefits from the richer context and more granular rubrics found in higher-level tasks, aligning with its success on the complex MMI dataset. This strong performance, achieved without any domain specific fine-tuning, demonstrates that the multi-agent, few-shot prompting strategy we developed for MMI scoring is not task-specific but can be a transferable methodology for multi-trait automated assessment.

\subsubsection{Ablation Study: The Impact of Examples and Rubrics}
\label{sec:results:ablation}
To further consolidate the contribution of our design choices, we conducted two ablation experiments on the ASAP dataset (results detailed in Table \ref{tab:asap_results}). First, we removed the 3-shot examples to test the necessity of in-context learning. Secondly, we replaced the detailed ASAP rubric with a generic, standardised rubric (mimicking the broad MMI rubric structure) to test the impact of rubric specificity. The full text of these reduced rubrics is provided in Appendix \ref{sec:appendix:asap_rubrics}.

The results confirm the necessity of our configuration. Removing the examples caused a decline in performance (e.g., Q8 Overall QWK dropped from 0.663 to 0.503), validating that 3-shot learning is essential for calibration even on standard AES tasks.

Similarly, using a reduced rubric also resulted in a performance drop compared to the base configuration (e.g., Q7 Overall QWK dropped from 0.638 to 0.561). This highlights two key findings: first, that detailed rubrics (like those in ASAP) naturally yield higher model agreement; and second, that our framework's strong performance on the MMI dataset is particularly notable given it operates under similar "generic rubric" constraints which we show here to be objectively harder to score.

\section{Discussion}
\label{sec:discussion}

This study set out to address the challenge of inter-rater reliability in MMI scoring. Our findings demonstrate that for a complex, context-dependent task like evaluating soft skills, a prompt engineering strategy outperforms common fine-tuning approaches. The performance of our multi-agent framework (Avg QWK 0.621) achieved nearly double the agreement of the best-performing fine-tuned model (Avg QWK 0.316).

The performance observed, where a task-specific prompt design with a powerful Large Language Model (LLM) exceeded the results of smaller, fine-tuned architectures, raises a critical discussion point for the field of applied NLP. Our two-stage framework successfully leverages the superior few-shot learning and complex reasoning capabilities of modern LLMs. This architecture demonstrates that for specific, reasoning-intensive tasks like multi-trait assessment, the correct prompt can effectively condition an LLM to achieve task performance that can outperform fine-tuning. As LLMs continue to improve, particularly in the 100B+ parameter category, we anticipate that specialised prompt engineering, using in-context examples, will likely surpass smaller fine-tuned models. 

The success of this methodology extends beyond MMI. The ablation study on the ASAP dataset confirmed the necessity of both few-shot examples and detailed rubrics, suggesting that our prompt-based design is a generalisable methodology for multi-trait assessment. Crucially, our results suggest this approach is particularly valuable in contexts where rubrics are sparse or limited. Where traditional AES-style fine-tuning struggled to adapt to the broad constraints of MMI scoring, our framework proved its ability to handle both abstract soft skills and conventional writing traits.

\section{Conclusion}
\label{sec:conclusion}

Interviews for competitive selection are often undermined by human scoring inconsistencies. This study addressed this challenge by investigating the application of Large Language Models to automated MMI scoring. We demonstrate that a simple multi-agent framework, which breaks down the evaluation into a preprocessing agent and specialised single-criterion scoring agents, outperforms smaller fine-tuned models. Our method, when implemented with a state-of-the-art model, achieves reliability on par with human experts and is generalisable to other domains. Most importantly, this work highlights that for complex soft-skill assessment, fine-tuning is not always necessary, presenting an effective methodology for enhancing the consistency, fairness, and efficiency of the MMI process.

\section*{Limitations}
\label{sec:Limitations}
While our findings are promising, we acknowledge several limitations. First, our prompt engineering methodology was developed exclusively on Llama 3.1 405B, which may have created a prompt structure "co-adapted" to their architecture, potentially disadvantaging other models in our SOTA comparison. Second, while our text-only approach was a deliberate choice to mitigate biases from video and audio, this creates a scoring mismatch. Our model is trained to predict ground-truth scores that human raters assigned while observing the full video, including non-verbal cues that are unavailable to our text-only system. Third, we did not conduct fine-tuning experiments using a highly detailed, question-specific rubric. This is because the human experts who generated the ground-truth scores assessed candidates using broad, abstract criteria rather than a granular checklist. Consequently, evaluating a model trained on a detailed rubric against these original human scores would yield an invalid QWK comparison. Furthermore, attempting to create a detailed rubric for these open-ended scenarios risks introducing new biases that do not align with the original assessors' judgments. This could distort the ground truth, potentially narrowing the range of valid interpretations that the MMI format is designed to accommodate. A further limitation is the current lack of explainability. While high inter-rater agreement is essential, a "black box" scorer remains a barrier to trust and adoption in critical domains. Our current implementation focuses on scoring accuracy, yet true decision support requires transparency. Future work must address this by prompting the multi-agent framework to generate justifications, referencing specific rubric criteria and extracting direct quotations. This evolution is necessary to transform the tool from a simple scorer into a transparent, evidence-based assistant for human evaluators. Finally, given that regulations in regions like the EU prohibit fully automated critical decision-making \citep{eu2024aiact}, the tool developed here is intended as a decision-support system to assist human interviewers, not to replace them.

\newpage

\bibliography{mmi_references} 

\clearpage
\appendix
\onecolumn 

\section{Scenario and Dataset Details}
\label{sec:appendix:scenarios}

\vspace{1em}

\subsection{MMI Scenario Summaries}
As the exact scenario-based questions cannot be shared, this section provides the summaries.
\vspace{1.5em}

\noindent\textbf{Question 3:} This scenario expects the candidate to describe how they would react in a social setting with a friend who abruptly departed in anger, triggered by the recent loss of a parent.
\vspace{1.5em}

\noindent\textbf{Question 4:} This scenario requires the candidate to explore their feelings of anxiety and inadequacy when facing a professional setback, particularly in comparison to a successful close friend.
\vspace{1.5em}

\noindent\textbf{Question 5:} This scenario expects the candidate to outline their response to a public ethical dilemma, requiring them to support a vulnerable person while navigating the negative behaviour of an impatient employee and other customers.
\vspace{1.5em}

\noindent\textbf{Question 6:} This question was delivered by audio where the candidate would listen to a conversation and must describe how they would respond to a situation where one member's refusal to continue due to distress clashes with another member's impatience, jeopardising the team's overall goal.
\vspace{2em}

\subsection{Descriptive Statistics}
Table \ref{tab:descriptive_stats} provides the descriptive statistics for candidate responses, referenced in Section~\ref{sec:task}. All scores are rounded to 3 significant figures.

\vspace{1.5em}

\begin{table*}[ht!]
\centering
\small
\begin{tabular}{l @{\hspace{1.5em}}rr@{\hspace{1.5em}} @{\hspace{1.5em}}rr@{\hspace{1.5em}} @{\hspace{1.5em}}rr@{\hspace{1.5em}} @{\hspace{1.5em}}rr@{}} 
\toprule
& \multicolumn{2}{c}{\textbf{Question 3}} & \multicolumn{2}{c}{\textbf{Question 4}} & \multicolumn{2}{c}{\textbf{Question 5}} & \multicolumn{2}{c}{\textbf{Question 6}} \\
\cmidrule(lr){2-3} \cmidrule(lr){4-5} \cmidrule(lr){6-7} \cmidrule(lr){8-9}
Criterion & Mean & Std Dev & Mean & Std Dev & Mean & Std Dev & Mean & Std Dev \\
\midrule
c2 & 5.12 & 1.04 & 4.96 & 1.12 & 5.13 & 0.968 & 5.20 & 0.998 \\
c3 & 5.10 & 0.980 & 5.06 & 1.05 & 5.14 & 0.930 & 5.19 & 0.933 \\
c4 & 5.34 & 1.01 & 4.95 & 1.09 & 4.94 & 1.00 & 5.02 & 1.09 \\
c5 & 5.09 & 1.00 & 4.89 & 1.13 & 5.29 & 0.962 & 5.30 & 1.04 \\
c6 & 5.16 & 1.08 & 5.02 & 1.18 & 5.09 & 1.03 & 5.09 & 1.09 \\
c7 & 5.36 & 0.959 & 5.11 & 1.09 & 5.02 & 0.999 & 4.89 & 1.13 \\
c8 & 5.10 & 0.996 & 4.99 & 1.10 & 5.16 & 0.993 & 4.89 & 1.23 \\
c9 & 5.24 & 0.992 & 5.23 & 1.14 & 5.32 & 0.983 & 4.95 & 1.15 \\
c10& 5.15 & 1.03 & 5.07 & 1.10 & 5.18 & 0.989 & 5.06 & 1.07 \\
\midrule
Score Avg. & 5.19 & 1.01 & 5.03 & 1.11 & 5.14 & 0.984 & 5.07 & 1.08 \\
\midrule
Word Length Avg. & \multicolumn{2}{c}{460} & \multicolumn{2}{c}{451} & \multicolumn{2}{c}{488} & \multicolumn{2}{c}{429} \\
\bottomrule
\end{tabular}
\caption{Descriptive statistics of candidate responses, including mean scores, standard deviation, and average word count per question. All scores rounded to 3 significant figures.}
\label{tab:descriptive_stats}
\end{table*}

\clearpage

\section{Prompt Templates}
\label{sec:appendix:prompts}

This appendix provides the core prompt templates used in our experimental methodology.

\vspace{2em}

\subsection{Template 1: Base (Zero-Shot) Prompt}
Used for baseline zero-shot experiments as described in Section~\ref{sec:methodology:prompting}.

\vspace{1.5em}
\noindent \textbf{System:} \\
You are an expert Multiple Mini Interviewer whose task is to score candidates' responses to a scenario-based question. Each response is scored against 9 criteria (c2 to c10), and each criterion is scored based on a 7-point Likert scale (7 being the highest).

\vspace{1em}
\noindent \textbf{User:} \\
The question asked was \verb|"{Question}"|
\vspace{0.5em}

\noindent For this specific scenario question, the criteria are the following: \\
\verb|"{List of all 9 criteria}"|

\vspace{0.5em}
\noindent Your task is to score the following response: \verb|"{candidate_response}"|

\vspace{0.5em}
\noindent Please provide the results in the following format and only provide the score, where x represents your chosen score: c2: x, c3: x, c4: x, c5: x, c6: x, c7: x, c8: x, c9: x, c10: x. Explanation is not needed.

\vspace{2em}

\subsection{Template 2: Few-Shot Prompt (Illustrating 1-Shot)}
This template illustrates the in-context learning strategy by showing how examples were added, as discussed in Section~\ref{sec:methodology:prompting}. Our optimal single-prompt configuration used this structure with 3 examples.

\vspace{1.5em}
\noindent \textbf{System:} \\
You are an expert Multiple Mini Interviewer whose task is to score candidates' responses to a scenario-based question. Each response is scored against 9 criteria (c2 to c10), and each criterion is scored based on a 7-point Likert scale (7 being the highest).

\vspace{1em}
\noindent \textbf{User:} \\
The question asked was:
\verb|"{Question}"|
\vspace{0.5em}

\noindent For this specific scenario question, the criteria are the following: \\
\verb|"{List of all 9 criteria}"|
\vspace{0.5em}

\noindent I will provide you with x examples with scores. The first example is \verb|"{Example 1}"|.
\vspace{0.5em}

\noindent The score received for the first example was c2: x, c3: x, c4: x, c5: x, c6: x, c7: x, c8: x, c9: x, c10: x.
\vspace{0.5em}

\noindent Your task is to score the following response: \verb|"{candidate_response}"|
\vspace{0.5em}

\noindent Please provide the results in the following format and only provide the score, where x represents your chosen score: c2: x, c3: x, c4: x, c5: x, c6: x, c7: x, c8: x, c9: x, c10: x. Explanation is not needed.

\vspace{1.5em}

\clearpage

\subsection{Template 3: Optimal Multi-Agent Framework}
The final, two-stage framework, as described in Section~\ref{sec:methodology:proposed}.

\vspace{1.5em}

\subsubsection*{Prompt 3a: The Preprocessing Agent}

\noindent \textbf{System:} \\
You are an expert in processing interview transcripts.

\vspace{1em}
\noindent \textbf{User:} \\
Given the following transcribed response, clean and refine it by:
\vspace{0.5em}
\begin{enumerate}[noitemsep,topsep=0pt,label=\arabic*)]
    \item Correcting minor transcription errors.
    \item Removing unnecessary filler words (e.g., "um", "uh").
    \item Removing conversational introductions. Omit opening phrases that are not part of the core answer, such as question repetitions, self identifications (e.g., "My name is...", "I am a..."), or acknowledgements of the question.
    \item Removing redundant statements (points repeated multiple times).
    \item Do not add content or improve the response.
    \item Your output must contain ONLY the cleaned structured transcript, no bullet points. Do not include any introductory phrases, explanations, or conversational text like "Here is the cleaned response:".
\end{enumerate}
\vspace{0.5em}

\noindent Original Response: \\
\verb|"{candidate_response}"|

\vspace{2em}

\subsubsection*{Prompt 3b: The Single-Criterion Scoring Agent}

\noindent \textbf{System:} \\
You are an expert interviewer evaluating a candidate's response in a Multiple Mini Interview (MMI).

\vspace{1em}
\noindent \textbf{User:} \\
The scenario question asked was: \\
\verb|"{Question}"|
\vspace{0.5em}

\noindent The response has already been preprocessed. Your task is to assign a score for the following criteria: \\
\verb|"{Criterion}"|
\vspace{0.5em}

\noindent Scoring Scale Descriptors: \\
\verb|"{Descriptors}"|
\vspace{0.5em}

\noindent Below area examples of responses with their score for this specific criterion:
\vspace{0.5em}

\noindent Example 1: \verb|"{Example 1 Text}"| \\
Score: cx: y
\vspace{0.5em}

\noindent Example 2: \verb|"{Example 2 Text }"| \\
Score: cx: y
\vspace{0.5em}

\noindent Example 3: \verb|"{Example 3 Text}"| \\
Score: cx: y
\vspace{0.5em}

\noindent Candidate Response: \\
\verb|"{preprocessed_candidate_response}"|
\vspace{0.5em}

\noindent Provide the score in the format: 'cx:y'. Where 'x' is the criterion number and 'y' is the score. Do not provide explanations for the score.

\clearpage

\section{Prompt Engineering and RAG Experiments}
\label{sec:appendix:prompt_rag_results}

This appendix presents the comprehensive results for the iterative development of prompting strategies (Section~\ref{sec:methodology:context_learning}) and Retrieval-Augmented Generation (RAG) experiments (Section~\ref{sec:methodology:RAG}). All data below utilises the Llama 3.1 405B Instruct model.

\subsection{In-Context Learning Strategies}
Table \ref{tab:prompt_engineering_results} details the performance metrics across all tested Zero-Shot and Few-Shot configurations. The results validate the selection of the 3-Shot (Low/Medium/High) strategy as the optimal configuration for this task.

\begin{table*}[ht!]
\centering
\small
\begin{tabular}{l l c c}
\toprule
\textbf{Method} & \textbf{Configuration} & \textbf{MSE$^\downarrow$} & \textbf{QWK$^\uparrow$} \\
\midrule
\textbf{0-Shot} & Base Prompt & 1.44 & 0.215 \\
\midrule
\multirow{3}{*}{\textbf{1-Shot}}
& Low Example & 1.25 & 0.278 \\
& Medium Example & 1.23 & 0.263 \\
& High Example & 1.22 & 0.257 \\
\midrule
\multirow{6}{*}{\textbf{2-Shot}}
& Low / Low & 1.24 & 0.316 \\
& Medium / Medium & 1.37 & 0.294 \\
& High / High & 1.35 & 0.262 \\
& Low / Medium & 1.27 & 0.310 \\
& Low / High & 1.23 & 0.311 \\
& Medium / High & 1.23 & 0.283 \\
\midrule
\multirow{10}{*}{\textbf{3-Shot}}
& Low / Low / Low & 1.18 & 0.329 \\
& Medium / Medium / Medium & 1.17 & 0.299 \\
& High / High / High & 1.27 & 0.280 \\
& Low / Low / Medium & 1.21 & 0.336 \\
& Low / Medium / Medium & 1.19 & 0.332 \\
& Low / Low / High & 1.19 & 0.325 \\
& \textbf{Low / Medium / High (Optimal)} & \textbf{1.17} & \textbf{0.363} \\
& Low / High / High & 1.22 & 0.313 \\
& Medium / Medium / High & 1.19 & 0.309 \\
& Medium / High / High & 1.25 & 0.298 \\
\midrule
\multirow{4}{*}{\textbf{4-Shot}}
& Low / Low / Low / Low & 1.23 & 0.274 \\
& Medium / Medium / Medium / Medium & 1.24 & 0.250 \\
& High / High / High / High & 1.45 & 0.212 \\
& Low / Low / Medium / High & 1.16 & 0.329 \\
\bottomrule
\end{tabular}
\caption{Impact of shot count and exemplar selection on scoring performance.}
\label{tab:prompt_engineering_results}
\end{table*}

\subsection{Retrieval-Augmented Generation (RAG) Strategies}
\label{sec:appendix:rag_details} 

For our retrieval implementation, we utilised the all-mpnet-base-v2 model from the Sentence-Transformers library to generate 768-dimensional dense vector representations of the transcripts. Similarity between the candidate response and potential in-context examples was calculated using Cosine Similarity.

Table \ref{tab:rag_results} compares standard RAG approaches against advanced calibration strategies. While the "Integrated Calibration" method adaptively improved performance, it did not surpass the static 3-Shot baseline established in Table \ref{tab:prompt_engineering_results}.

\begin{table*}[ht!]
\centering
\small
\begin{tabular}{l l c c}
\toprule
\textbf{RAG Strategy} & \textbf{Description} & \textbf{MSE$^\downarrow$} & \textbf{QWK$^\uparrow$} \\
\midrule
\textbf{RAG 1-Shot} & Retrieve 1 most similar response & 1.28 & 0.282 \\
\textbf{RAG 2-Shot} & Retrieve 2 most similar responses & 1.24 & 0.285 \\
\textbf{RAG 3-Shot} & Retrieve 3 most similar responses & 1.25 & 0.292 \\
\textbf{RAG 4-Shot} & Retrieve 4 most similar responses & 1.22 & 0.298 \\
\midrule
\textbf{RAG Calib. + 1-Shot} & 3-Shot Static Base + 1 Retrieved Example & 1.19 & 0.305 \\
\textbf{RAG Integrated Calib.} & \textbf{Adaptive Replacement of Base Exemplars} & \textbf{1.18} & \textbf{0.329} \\
\bottomrule
\end{tabular}
\caption{Performance comparison of RAG strategies.}
\label{tab:rag_results}
\end{table*}

\section{Fine-Tuning Setup Details}
\label{sec:appendix:finetune_setup}

This appendix provides the implementation details for the fine-tuning experiments described in Section \ref{sec:results:finetuning_baseline}.

\vspace{0.5em}

We divided the dataset into an 80 / 20 train–test split. To manage the computational demands of fine-tuning the Llama 3.1 8B model, we employed several optimisation techniques. Training was facilitated by the Unsloth library for memory-efficient fine-tuning, using a 4-bit quantised model to remain within the memory limits of a single T4 GPU.

\vspace{0.5em}

We applied Low-Rank Adaptation (LoRA), a parameter-efficient fine-tuning method, with the following configuration: rank ($r$) = 16, $\alpha$ = 32, dropout = 0.05. Adaptation was applied to the key attention and feed-forward network modules. The model was trained for 3 epochs using the adamw\_8bit optimiser and a learning rate of 2e-4.

\vspace{0.5em}

All experiments were performed using the same training and evaluation protocol to ensure fair comparison between architectures and configurations.

\vspace{0.55em}

\section{Error Distribution Analysis}
\label{sec:appendix:error_dist}

\begin{figure}[!hb] 
    \centering
    \includegraphics[width=\textwidth, height=0.45\textheight, keepaspectratio]{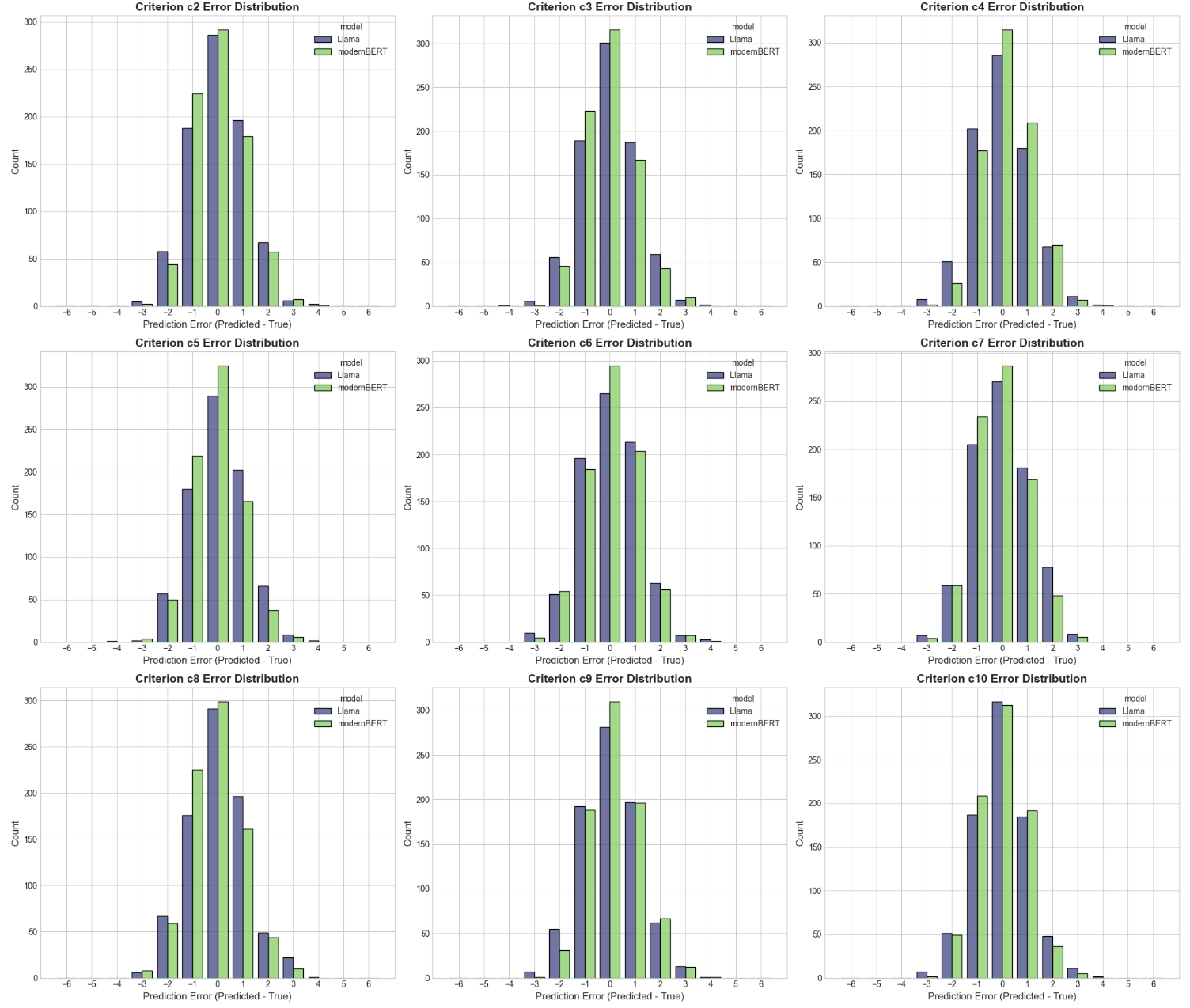} 
    \caption{Prediction Error Distribution (Predicted -- True) for RMTS-adapted Llama 3.1 8B and modernBERT across all nine criteria.}
    \label{fig:error_dist}
\end{figure}

This appendix provides the full prediction error distribution analysis referenced in Section~\ref{sec:results:analysis_finetuning}. Figure \ref{fig:error_dist} visualises the distribution of $(\text{Predicted} - \text{True})$ scores for both the RMTS-adapted Llama 3.1 8B and modernBERT models across all nine criteria.

\vspace{0.5em}

The balanced distribution around zero suggests that the models do not exhibit a systematic bias (e.g., consistent overestimation or underestimation), while the wide spread indicates high variance and inaccuracy in individual predictions.

\vspace{1.5em}

\clearpage

\section{Detailed SOTA Model Performance}
\label{sec:appendix:sota_details}

This appendix provides the full, criterion-by-criterion performance breakdown for the SOTA model comparison, as referenced in Section \ref{sec:results:sota}. Table \ref{tab:appendix:sota_mse} shows the Mean Squared Error, and Table \ref{tab:appendix:sota_qwk} shows the Quadratic Weighted Kappa.

\vspace{2em}

\begin{table*}[ht!]
\centering
\small
\begin{tabular}{l l cccc}
\toprule
Question & Attribute & Llama 4 Maverick & GPT 5 & Gemini 2.5 Pro & DeepSeek Reasoner \\
\midrule
\multirow{10}{*}{3} & c2 & 0.873 & 0.793 & 0.982 & 1.13 \\
& c3 & 0.681 & 1.16 & 1.18 & 0.776 \\
& c4 & 0.704 & 0.956 & 1.08 & 0.953 \\
& c5 & 0.648 & 1.08 & 1.09 & 0.756 \\
& c6 & 0.751 & 0.921 & 0.991 & 0.706 \\
& c7 & 0.695 & 0.915 & 1.28 & 0.820 \\
& c8 & 0.704 & 0.915 & 1.13 & 0.625 \\
& c9 & 0.683 & 1.11 & 1.25 & 0.857 \\
& c10 & 0.663 & 1.01 & 1.04 & 0.663 \\
\cmidrule{2-6}
& Average & 0.711 & 0.984 & 1.11 & 0.810 \\
\midrule
\multirow{10}{*}{4} & c2 & 1.07 & 0.947 & 1.14 & 1.07 \\
& c3 & 0.785 & 1.09 & 1.34 & 0.729 \\
& c4 & 0.735 & 1.35 & 1.42 & 1.01 \\
& c5 & 0.667 & 1.86 & 2.21 & 1.10 \\
& c6 & 0.732 & 0.909 & 1.17 & 0.749 \\
& c7 & 0.761 & 0.764 & 0.920 & 0.746 \\
& c8 & 0.723 & 0.915 & 1.41 & 0.702 \\
& c9 & 0.856 & 0.912 & 1.50 & 1.12 \\
& c10 & 0.702 & 0.932 & 0.947 & 0.776 \\
\cmidrule{2-6}
& Average & 0.781 & 1.08 & 1.34 & 0.889 \\
\midrule
\multirow{10}{*}{5} & c2 & 1.40 & 1.12 & 1.64 & 1.87 \\
& c3 & 1.23 & 1.69 & 2.16 & 1.43 \\
& c4 & 1.13 & 1.34 & 1.76 & 1.43 \\
& c5 & 1.31 & 2.87 & 1.57 & 1.35 \\
& c6 & 1.21 & 1.50 & 1.81 & 1.23 \\
& c7 & 0.991 & 1.47 & 1.59 & 1.14 \\
& c8 & 1.27 & 2.20 & 1.58 & 1.34 \\
& c9 & 1.01 & 1.35 & 1.16 & 1.16 \\
& c10 & 1.13 & 1.65 & 1.63 & 1.33 \\
\cmidrule{2-6}
& Average & 1.19 & 1.69 & 1.65 & 1.36 \\
\midrule
\multirow{10}{*}{6} & c2 & 0.708 & 1.04 & 0.831 & 1.19 \\
& c3 & 0.746 & 1.45 & 1.10 & 0.957 \\
& c4 & 0.847 & 1.57 & 0.866 & 1.12 \\
& c5 & 0.850 & 1.14 & 1.08 & 1.02 \\
& c6 & 0.787 & 0.951 & 0.913 & 0.778 \\
& c7 & 0.743 & 1.21 & 0.822 & 0.743 \\
& c8 & 1.01 & 1.20 & 0.988 & 0.992 \\
& c9 & 0.720 & 1.35 & 1.05 & 1.20 \\
& c10 & 0.836 & 1.32 & 1.33 & 0.882 \\
\cmidrule{2-6}
& Average & 0.805 & 1.25 & 0.997 & 0.986 \\
\bottomrule
\end{tabular}
\caption{Detailed Criterion-Level Performance of SOTA Models for Mean Squared Error.}
\label{tab:appendix:sota_mse}
\end{table*}

\clearpage

\begin{table*}[ht!]
\centering
\small
\begin{tabular}{l l cccc}
\toprule
Question & Attribute & Llama 4 Maverick & GPT 5 & Gemini 2.5 Pro & DeepSeek Reasoner \\
\midrule
\multirow{10}{*}{3} & c2 & 0.583 & 0.370 & 0.709 & 0.314 \\
& c3 & 0.576 & 0.269 & 0.641 & 0.345 \\
& c4 & 0.629 & 0.462 & 0.630 & 0.304 \\
& c5 & 0.744 & 0.494 & 0.745 & 0.521 \\
& c6 & 0.700 & 0.565 & 0.806 & 0.535 \\
& c7 & 0.653 & 0.511 & 0.653 & 0.574 \\
& c8 & 0.626 & 0.498 & 0.646 & 0.617 \\
& c9 & 0.627 & 0.399 & 0.594 & 0.478 \\
& c10 & 0.714 & 0.460 & 0.804 & 0.568 \\
\cmidrule{2-6}
& Average & 0.650 & 0.448 & 0.692 & 0.473 \\
\midrule
\multirow{10}{*}{4} & c2 & 0.610 & 0.398 & 0.544 & 0.379 \\
& c3 & 0.653 & 0.282 & 0.478 & 0.528 \\
& c4 & 0.665 & 0.371 & 0.506 & 0.417 \\
& c5 & 0.723 & 0.298 & 0.435 & 0.457 \\
& c6 & 0.748 & 0.606 & 0.695 & 0.704 \\
& c7 & 0.690 & 0.696 & 0.682 & 0.626 \\
& c8 & 0.737 & 0.606 & 0.488 & 0.678 \\
& c9 & 0.665 & 0.494 & 0.408 & 0.529 \\
& c10 & 0.697 & 0.502 & 0.731 & 0.577 \\
\cmidrule{2-6}
& Average & 0.688 & 0.473 & 0.552 & 0.544 \\
\midrule
\multirow{10}{*}{5} & c2 & 0.482 & 0.413 & 0.623 & 0.346 \\
& c3 & 0.292 & 0.148 & 0.210 & 0.106 \\
& c4 & 0.426 & 0.572 & 0.560 & 0.423 \\
& c5 & 0.345 & 0.151 & 0.299 & 0.212 \\
& c6 & 0.594 & 0.460 & 0.457 & 0.378 \\
& c7 & 0.485 & 0.560 & 0.445 & 0.291 \\
& c8 & 0.417 & 0.218 & 0.488 & 0.394 \\
& c9 & 0.363 & 0.333 & 0.629 & 0.512 \\
& c10 & 0.497 & 0.299 & 0.550 & 0.276 \\
\cmidrule{2-6}
& Average & 0.434 & 0.351 & 0.473 & 0.327 \\
\midrule
\multirow{10}{*}{6} & c2 & 0.707 & 0.485 & 0.745 & 0.256 \\
& c3 & 0.602 & 0.293 & 0.581 & 0.329 \\
& c4 & 0.697 & 0.268 & 0.825 & 0.488 \\
& c5 & 0.755 & 0.590 & 0.700 & 0.544 \\
& c6 & 0.750 & 0.636 & 0.792 & 0.637 \\
& c7 & 0.715 & 0.393 & 0.835 & 0.667 \\
& c8 & 0.619 & 0.494 & 0.779 & 0.587 \\
& c9 & 0.881 & 0.556 & 0.849 & 0.581 \\
& c10 & 0.676 & 0.407 & 0.686 & 0.594 \\
\cmidrule{2-6}
& Average & 0.711 & 0.458 & 0.754 & 0.520 \\
\bottomrule
\end{tabular}
\caption{Detailed Criterion-Level Performance of SOTA Models for Quadratic Weighted Kappa.}
\label{tab:appendix:sota_qwk}
\end{table*}


\section{ASAP Generic Rubrics}
\label{sec:appendix:asap_rubrics}

To test the impact of rubric specificity in Section \ref{sec:results:ablation}, we replaced the detailed, domain-specific ASAP rubrics with the following generic scoring scales. These were designed to imitate the broad, abstract nature of the MMI rubric used in our main dataset.

\vspace{1em}

\subsection{Reduced Rubric for Question 7 (Scale 0-3)}
\begin{itemize}[noitemsep,topsep=0pt]
    \item \textbf{(3) High:} The response demonstrates a clear and consistent command of this trait.
    \item \textbf{(2) Medium:} The response demonstrates a reasonable command of this trait, though there may be some errors or lack of depth.
    \item \textbf{(1) Low:} The response demonstrates a weak command of this trait with significant errors or lack of understanding.
    \item \textbf{(0) Unsatisfactory:} The response is totally irrelevant, illegible, or shows no evidence of this trait.
\end{itemize}

\vspace{0.5em}

\subsection{Reduced Rubric for Question 8 (Scale 1-6)}
\begin{itemize}[noitemsep,topsep=0pt]
    \item \textbf{(6) Excellent:} Exceptional performance. The response is highly insightful, skillful, and flawless in its execution of this trait.
    \item \textbf{(5) Strong:} Very good performance. The response shows command of the trait with only minor imperfections.
    \item \textbf{(4) Competent:} Satisfactory performance. The response is adequate and meets expectations for this trait, though it lacks depth or polish.
    \item \textbf{(3) Developing:} Inconsistent performance. The response shows some understanding of this trait but is uneven or has noticeable weaknesses.
    \item \textbf{(2) Limited:} Weak performance. The response struggles significantly with this trait and errors impede understanding.
    \item \textbf{(1) Unsatisfactory:} Poor performance. The response fails to demonstrate this trait effectively or is incoherent.
\end{itemize}

\end{document}